\documentclass[10pt,twocolumn,letterpaper]{article}

\usepackage[pagenumbers]{cvpr} 

\usepackage{booktabs,multirow}
\usepackage[table]{xcolor}
\usepackage{siunitx}

\usepackage{array} 
\newcolumntype{Z}{@{\hspace{2pt}}}
\usepackage{booktabs}
\usepackage{tabularx}
\usepackage{pifont} 

\usepackage{siunitx}
\usepackage{graphicx}
\usepackage{subcaption}
\usepackage{caption}
\usepackage{float}
\usepackage{amsmath}
\usepackage{mathtools}
\DeclareMathOperator{\softmax}{softmax}

\definecolor{cvprblue}{rgb}{0.21,0.49,0.74}
\usepackage[pagebackref,breaklinks,colorlinks,allcolors=cvprblue]{hyperref}

\usepackage{xspace}

\newcommand{\name}{\textit{{I-Scene}\xspace}}

\newcommand{\tb}[1]{\textbf{#1}}

\title{\name: 3D Instance Models are Implicit Generalizable Spatial Learners}

\author{Lu Ling$^{1}$\thanks{Corresponding author.},
Yunhao Ge$^2$,
Yichen Sheng$^2$,
Aniket Bera$^{1}$,
\\
$^1$Purdue University \quad 
$^2$NVIDIA Research
\\
\\
\href{https://luling06.github.io/I-Scene-project/}{\textit{I-Scene} Project Page}
}

\begin{document}

\twocolumn[{
    \renewcommand\twocolumn[1][]{#1}
    \maketitle
    \begin{center}
        \begin{minipage}{\linewidth}\centering
  \includegraphics[width=\linewidth]{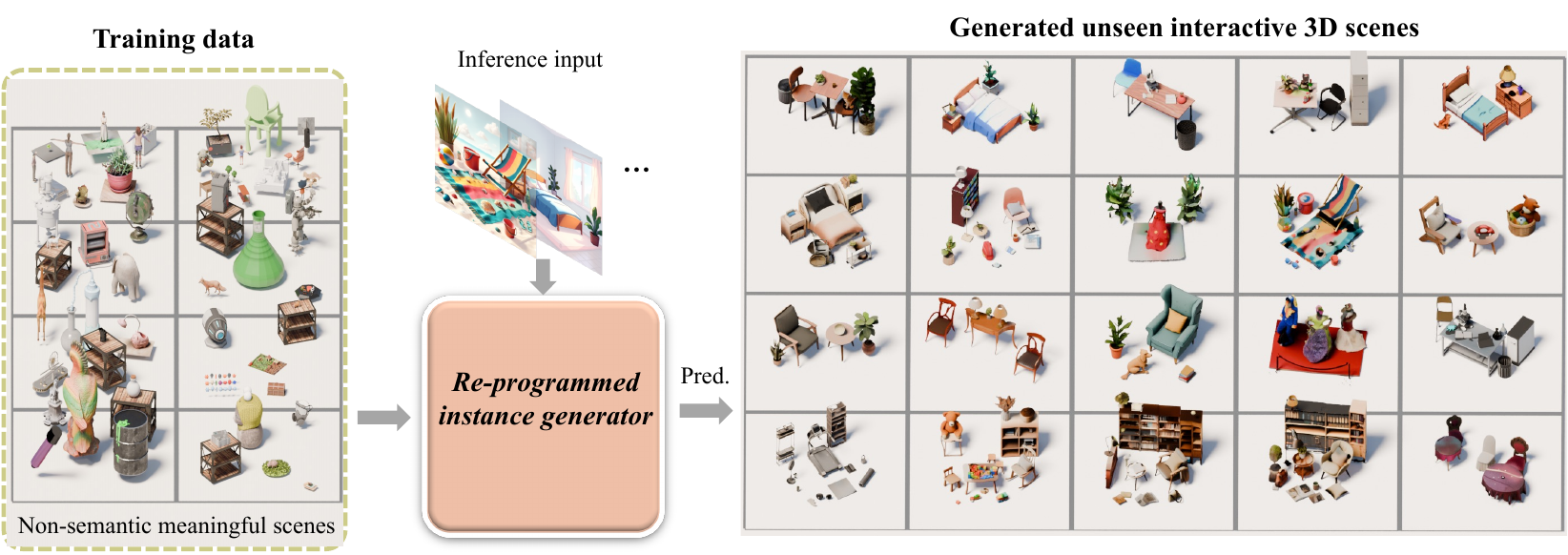}
  \captionof{figure}{A pre-trained 3D instance model is re-programmed into a scene-level spatial learner. 
  It learns spatial priors from non-semantic scenes (randomly composed instances) in a feed-forward manner, producing coherent layouts and unseen interactive 3D scenes.
  }
  \label{fig:teaser}
\end{minipage}

    \end{center}
}]

\begingroup
\renewcommand\thefootnote{\fnsymbol{footnote}}
\footnotetext[1]{Corresponding author.}
\endgroup

\begin{abstract}

Generalization remains the central challenge for interactive 3D scene generation. 
Existing learning‑based approaches ground spatial understanding in limited scene dataset, restricting generalization to new layouts.
We instead reprogram a pre‑trained 3D instance generator to act as a scene‑level learner, replacing dataset-bounded supervision with model-centric spatial supervision.
This reprogramming unlocks the generator's transferable spatial knowledge, enabling generalization to unseen layouts and novel object compositions.
Remarkably, spatial reasoning still emerges even when the training scenes are randomly composed objects. 
This demonstrates that the generator’s transferable scene prior provides a rich learning signal for inferring proximity, support, and symmetry from purely geometric cues.
Replacing widely used canonical space, we instantiate this insight with a view‑centric formulation of the scene space, yielding a fully feed‑forward, generalizable scene generator that learns spatial relations directly from the instance model.
Quantitative and qualitative results show that a 3D instance generator is an implicit spatial learner and reasoner, pointing toward foundation models for interactive 3D scene understanding and generation. The code is accessible at  \href{https://luling06.github.io/I-Scene-project/}{\textcolor{red}{the project page}}.

\end{abstract}    
\section{Introduction}
\label{sec:intro}
Generalization remains the central obstacle to interactive 3D scene generation. An effective system must produce editable, affordance‑aware, and spatially coherent object arrangements, capabilities critical for virtual content creation, simulation, and embodied AI.
Recent end‑to‑end approaches extend powerful image‑to‑3D \emph{instance} priors \cite{xiang2025structured,tochilkin2024triposr} to multiple objects and relations in a single pass, yet their spatial understanding is typically learned from curated scene datasets whose limited coverage does not scale and ultimately constrains generalization to new layouts.

Recent methods learn layouts directly from annotated scenes \cite{yang2024physcene,tang2024diffuscene,meng2025scenegen,huang2025midi}; for example, SceneGen \cite{meng2025scenegen} explicitly models poses on 3D‑FRONT \cite{fu20213d}, and MIDI \cite{huang2025midi} learns inter‑object relations from those annotations.
However, available interactive scene datasets are limited in scale, diversity, and spatial variety.
For instance, the widely used interactive 3D scene dataset 3D\mbox{-}FRONT contains only $\sim$20K indoor bedroom and living\mbox{-}room scenes and under\mbox{-}represents small and supporting objects.
Consequently, scene layout‑supervised models often \emph{overfit to dataset‑specific biases} and fail to generalize to broader scene distributions—for example, small objects placed on or behind large furniture or outdoor settings.

Our key insight is that a pre‑trained 3D instance model implicitly encodes transferable spatial knowledge—depth, occlusion, scale, and support—even though it outputs single‑mesh geometry.
We \emph{reprogram} this prior as a scene-level spatial learner to provide model-centric spatial supervision and enables it generalize to unseen layouts, eliminating dependence on curated scene annotations.

A key obstacle to turning an instance generator into a scene-level learner is that its canonical object space suppresses spatial sensitivity: different views are collapsed into the same canonical representation, destroying layout cues needed for scene reasoning. 
To overcome this limitation, we replace the widely used canonical space with a view-centric scene space, in which each scene is represented in a view-dependent coordinate frame that preserves the layout relationship between the camera view and the scene.

Reprogramming the instance model to a spatial learner and training it in the view-centric space yields a fully feed‑forward formulation of interactive scene generator that learns spatial relations directly from instance prior.
Via this formulation, training entirely on non-semantic synthetic scenes in which objects are randomly composed without meaningful relations surprisingly yields to strong spatial reasoning ability, surpassing dataset-bounded baselines and  robustly generalizes to various layouts.

These results indicate that non-semantic synthetic scenes are sufficient for spatial learning and suggest a promising scaling path for interactive 3D scene understanding and generation.
Our contributions are summarized as follows:
\begin{itemize}
    \item \textbf{Model‑centric supervision.} We reprogram a pre‑trained 3D instance prior to function as a scene‑level spatial learner, revealing its transferable spatial knowledge without relying on scene-level annotations.
    \item \textbf{View‑centric scene space.} We replace the canonical object space with a view-aligned shared scene representation that preserves geometric and relational cues, enabling a fully feed-forward scene generator.
    \item \textbf{Dataset‑independent layout learning.} We show that the reprogrammed instance prior's strong spatial learning capability from  non‑semantic, randomly composed synthetic scenes, relaxing annotated data dependency.
    \item \textbf{Strong Generalization.} Despite being trained solely on random layouts, our model surpasses SOTA methods trained on 3D‑FRONT and transfers robustly to diverse unseen layouts.
\end{itemize}

\section{Related Work}

\noindent \textbf{Single geometry 3D scene generation.} 
One line of work represents entire scenes as a single global structure
guided by text or visual prior.
For example, Wonderland~\cite{liang2025wonderland} predicts 3D Gaussian Splatting in a feed-forward manner from a single image; WonderWorld~\cite{yu2025wonderworld} generates connected scenes from a single view using layered Gaussian surfels. 
Similarly, WorldExporer~\cite{schneider_hoellein_2025_worldexplorer} 

targets text-driven, 3D-consistent but non-disentangled scenes. 
These monolithic approaches achieve impressive view consistency and speed, but lack instance accordance, limiting fine-grained editing, instance interaction, and physics reasoning.

\medskip
\noindent \textbf{Compositional 3D interactive scene generation.}
Recent pipelines decompose scenes into perception and assembly: instance detection/segmentation and depth provide amodal cues, then objects are retrieved or generated, finally are assembled into a scene.
\emph{Gen3DSR}~\cite{ardelean2024gen3dsr} follows a modular divide-and-conquer strategy;
\emph{Deep Prior Assembly}~\cite{zhou2024zero} integrates frozen priors (e.g., SAM~\cite{kirillov2023segment}, diffusion, Shap-E~\cite{jun2023shap}) with layout fitting;
\emph{REPARO}~\cite{han2024reparo} performs compositional asset generation with differentiable 3D layout alignment;
\emph{CAST}~\cite{yao2025cast} applies component-aware analysis with SDF-based physics correction.
On the text-driven side, graph/LLM planners (e.g., \emph{LayoutGPT}~\cite{feng2023layoutgpt}, \emph{Holodeck}~\cite{yang2024holodeck}, \emph{GALA3D}~\cite{zhou2024gala3d}) synthesize scene graphs or relational constraints that downstream stages realize with assets and solvers.
\emph{Scenethesis}~\cite{ling2025scenethesis} and \emph{SceneWeaver}~\cite{yang2025sceneweaver} combines LLM planning, vision guidance, and physics-aware optimization with a plausibility judge.
These approaches provide high fidelity and open-ended semantics, but are sensitive to early perception/planning errors, potential retrieval/solver mismatch, and incur per-scene optimization, limiting throughput and scalability.

\noindent \textbf{Learning-based 3D interactive scene generation.}
Most recent approaches learn spatial layout distributions from curated 3D scene datasets~\cite{fu20213d}, differing mainly in how they instantiate object assets.
i) \textbf{Layout-first scene synthesis.} 
model a room as an unordered set of object tokens with attributes such as category, size, pose, which are later instantiated as geometry via retrieval or generation. 
For instance, \emph{ATISS~\cite{paschalidou2021atiss}}, \emph{SceneFormer~\cite{wang2021sceneformer}}) 
\emph{DiffuScene~\cite{tang2024diffuscene}}, \emph{MiDiffusion~\cite{hu2024mixed}}, \emph{DeBaRA~\cite{maillard2024debara}}, and  \emph{PhyScene~\cite{yang2024physcene}} condition on room type or floor plan, enabling controllable scene completion and synthesis;
ii) \textbf{End-to-end multi-instance method.}
With the advancement of pre-trained image-to-3D object model, recent approaches jointly model multiple objects and their spatial relations, reducing cascade errors and avoiding retrieval/solver loops at test time.
\emph{MIDI-3D ~\cite{huang2025midi}} and \emph{SceneGen~\cite{meng2025scenegen}} model multiple assets and their implicit inter-object relations or instance explicit poses in one feed-forward pass.
\emph{PartCrafter~\cite{lin2025partcrafter}} extends compositional latent diffusion transformers to jointly denoise parts/objects into explicit triangle meshes. 
Both paradigms achieve strong control and coherence but remain constrained by the biases and limited diversity of existing scene datasets, often struggling with rare layouts or complex arrangements such as small objects supported by or occluded behind larger ones.

Our method uses the same input protocol as prior learning-based approaches but differs fundamentally in design.
We introduce a feed-forward, view-centric approach that jointly reasons over global scene context and instance generation.
This design enables (i) layout generalization beyond dataset biases, capturing richer spatial relations such as small/supporting objects; and (ii) end-to-end inference without retrieval or solver handoffs, avoiding the stage-wise errors and latency of compositional pipelines.
Unlike prior methods constrained by curated datasets or optimization-heavy reasoning, our framework could learn spatial knowledge directly from non-semantic synthetic layouts, achieving dataset independence while preserving instance-level editability.

\begin{figure*}[t]
    \centering
    \includegraphics[width=\linewidth]{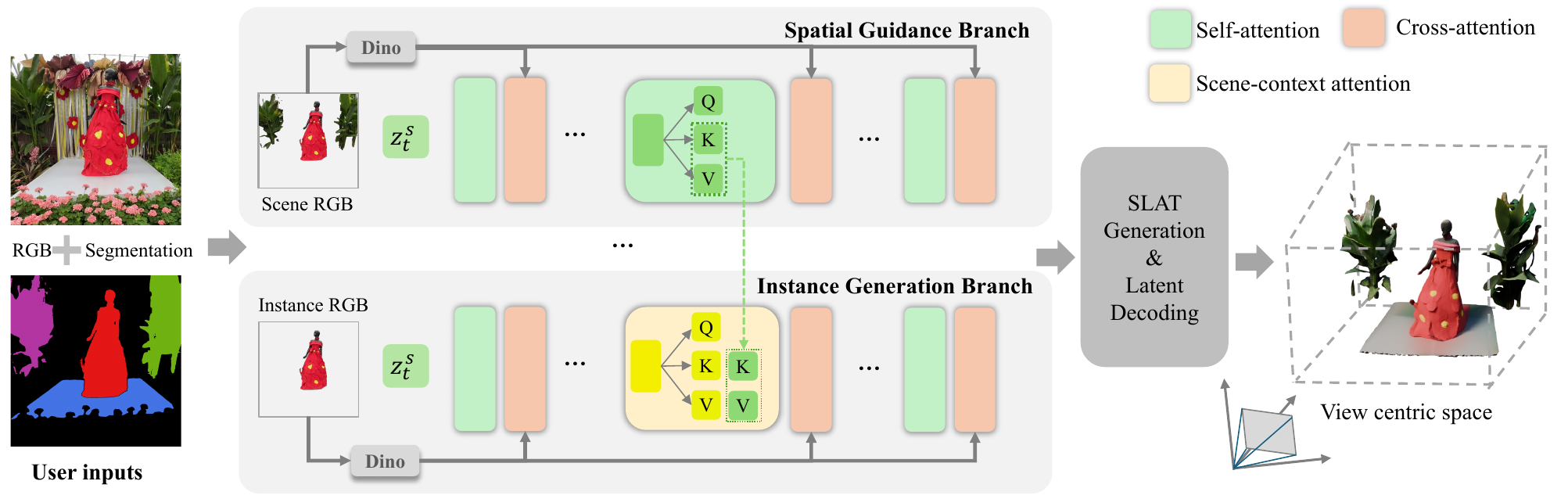}
    \caption{\textbf{Overview}. \name\ has two branches: (i) spatial guidance branch takes scene RGB as input and provides spatial anchor for each instance generation. (2) Instance branch takes instance RGB and scene context tokens and output the instance in view centric space. }
    \label{fig:placeholder}
\end{figure*}

\label{subsection:method_overview}
\section{Method}

\subsection{Problem Definition}
\label{subsection:problem_definition}
Given a single image $I_{\text{scene}} \in \mathbb{R}^{H \times W \times 3}$ and its instance masks $\{m_i\}_{i=1}^N$, \name\ generates a set of independently manipulable 3D instances $\mathcal{A}=\{A_i\}_{i=1}^N$, coherently placed in the scene space such that the spatial layout aligns with the input image.

\subsection{Overview}
~\autoref{fig:placeholder} presents the overview of our generalizable spatial learner(\name). \name\ reprograms an image-to-3D instance foundation model to a spatial leaner. 
We use TRELLIS~\cite{xiang2025structured} backbone in the experiment. 
\name\ only modifies the sparse structure transformer. 
Other stages are kept the same. 
\name\ has two branches: \textbf{spatial guidance branch} and \textbf{instance generation branch}. 
The two branches share the same weight and are trained jointly. 

\textbf{Spatial guidance branch.} 
The spatial guidance branch takes the scene RGB image as input and predicts the scene as a single geometry represented as a sparse set of active voxels. Following TRELLIS, the branch outputs
\begin{equation}
    f_{\text{scene}} = \{(f_i, p_i)\}_{i=1}^{L},
\end{equation}
where $p_i$ denotes the position of the $i$-th active voxel, $f_i$ is its corresponding local feature, and $L$ is the number of active voxels.
This branch provides two key functions:
(i) it produces a global scene layout that guides instance-level generation; and
(ii) it establishes a shared scene coordinate frame (an “anchor axis”) that all instances reference.
Without this global anchor, instances would be generated independently and the resulting composited scene layout would become incoherent.

\paragraph{Instance generation branch.}
The instance generation branch takes an instance RGB image $I_{\text{inst}}$ as input, conditioned on the spatial guidance latent $z_{\text{scene}}$, and learns a function $F$ to predict voxelized instance features:
\begin{equation}
    f_{\text{inst}} = F\big(I_{\text{inst}}, z_{\text{scene}}\big),
\end{equation}

The instance generation not only learns the instance local geometry, but also its pose in the the scene. 
As $z_{\text{scene}}$ already provides the scene layout, $F$ just needs to focus on geometry generation and follow the scene layout guidance. 
This effectively converts the instance generator from implicitly encoding scene priors to explicitly learning spatial layout.
This key formulation unlocks generalizable 3D scene generation.
In detail, $F$ is achieved by scene-context attention.

\subsection{Scene-Context Attention}
\label{sec:scenecontext}
We not only need to learn $F$, but also need to preserve the base model's prior as we do not want $z_{scene}$ to be catastrophically forgotten. 
So we need to minimize the change to the base model. 
We present scene-context attention to achieve this goal.

We transform some of the original self-attention layers to scene-context attention(SCA) layers.  
In the original self-attention layer, let $(Q_i, K_i, V_i)$ be queries/keys/values from instance $i$ and $(Q_s, K_s, V_s)$ be the queries/keys/values from the spatial guidance branch. 

\begin{equation}
    \tilde{K}_i = [K_i; \, K_{\text{s}}], \quad 
    \tilde{V}_i = [V_i; \, V_{\text{s}}],
\end{equation}
\begin{equation}
    \text{SCA}(Q_i, \tilde{K}_i, \tilde{V}_i)
    = \text{softmax}\!\left(\frac{Q_i \tilde{K}_i^\top}{\sqrt{d}}\right) \tilde{V}_i,
\end{equation}
Intuitively, the instance generation is not only based on its own $K_i$ and $V_i$, but also conditions on $K_{s}$ and $V_{s}$. 
SCA is a natural modification to the backbone as it does not change the latent distribution, making minimal changes to the prior. 
An extreme example can demonstrate this property: when the $K_i/V_i$ is the same with the scene $K_{s}/V_{s}$, meaning when the scene and instance input is exact the same, the SCA gives equivalent result of self-attention layers.
We show the mathematical proof in the \textit{Appendix}. 

\begin{figure}[t]
  \centering
  \begin{subfigure}[b]{0.65\linewidth}
    \centering
    \includegraphics[width=\linewidth]{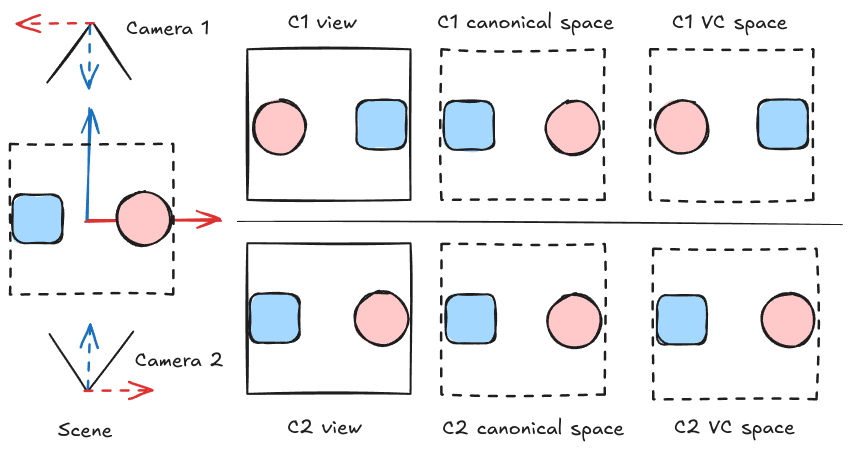}
    \caption{View-centric space}
    \label{fig:view-centric}
  \end{subfigure}\hfill
  \begin{subfigure}[b]{0.34\linewidth}
    \centering
    \includegraphics[width=\linewidth]{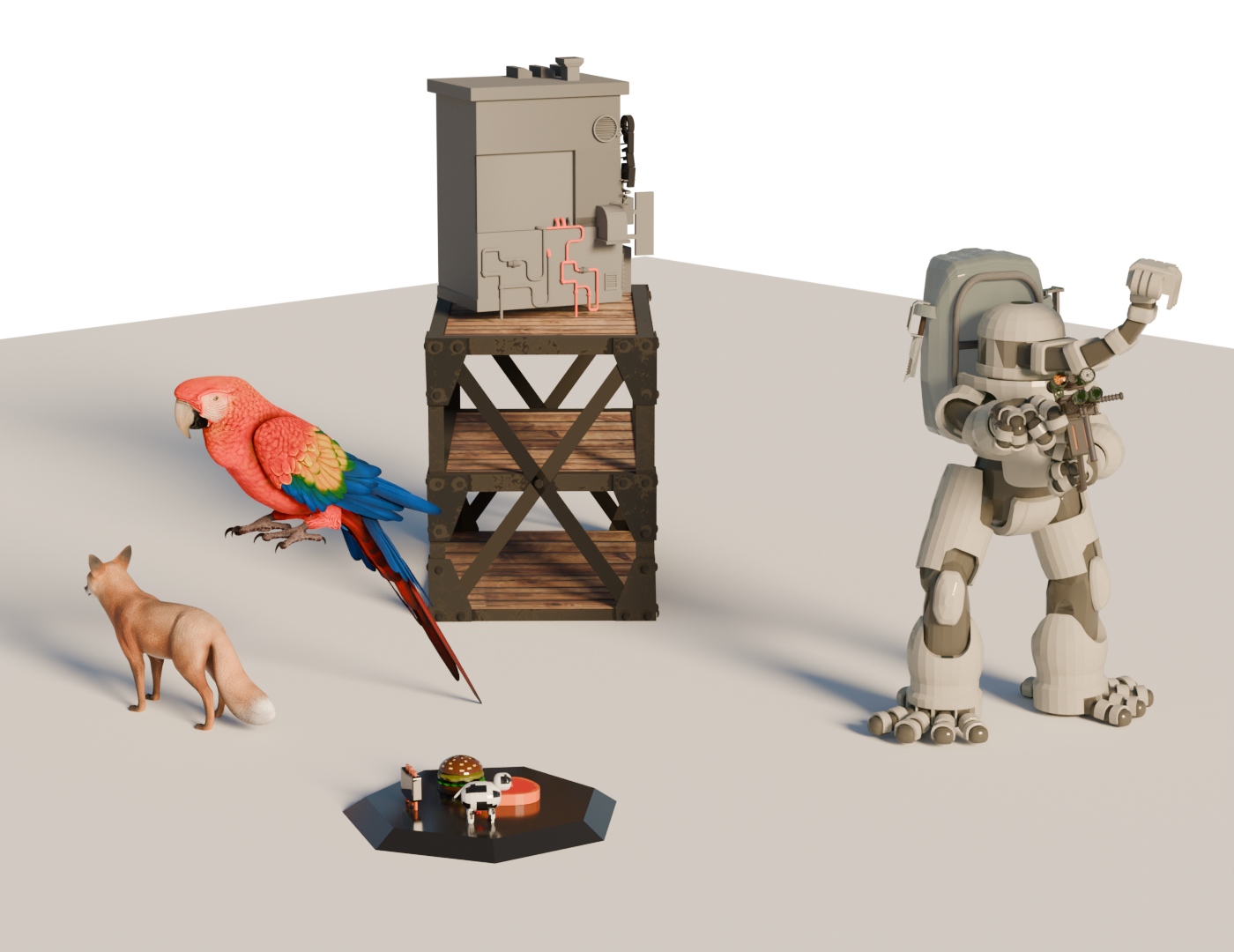}
    \caption{Non-Semantic scene}
    \label{fig:random-scene}
  \end{subfigure}
  \caption{(a).Canonical space scene is view invariant. View centric(VC) space is view-dependent, which encodes strict spatial relationship between the image space and scene space. (b) a non-semantic random 3D scene example. Objects have random poses and are collision-free.}
  \label{fig:vc-and-non-semantic-scene}
\end{figure}

\subsection{View-Centric(VC) Space}
Existing relevant works~\cite{huang2025midi, meng2025scenegen} follow image-to-3D base model and use canonical space for scene representation.  
We observe using canonical space limits the learning of $F$ and hurts the model generalization ability.
As illustrated in \autoref{fig:vc-and-non-semantic-scene} (a), given two camera ($C1$ and $C2$) and the scene composed of a blue square and a red circle in the space, the camera views of the scene are different as they are in different location and orientation, but the instance in the scene stay at the same position in the canonical space. 
So this formulation makes the $F$ ignores the object spatial position in the view image $I$ and only focuses on the local shape of each instance in the image. 
When training on some indoor dataset like 3D-FRONT, it is not significantly hurting $F$ as there are limited object in the scene and each instance usually has very different shapes. 
But if the scene has identical objects, e.g. several identical chairs having similar poses, $F$ often put duplicated chairs at the same position.
We argue that the object spatial layout in the RGB image is also important and provides strong hint on its 3D layout pose.  
So we propose view-centric space where the space axis is based on the camera pose. 
As shown in \autoref{fig:vc-and-non-semantic-scene} (a), given the same two camera setting, the spatial layout of the two objects follows the change of the camera pose coherently. 
When training \name\ in VC space, $F$ achieves much better generalization results as shown in our ablation study(\autoref{tab:ablation}).

\subsection{Non-Semantic Synthetic 3D Scene}\label{sec:non-semantic synthetic scene}
3D-FRONT is a domain specific dataset that only has limited indoor assets. 
Although with our SCA and VC space, \name\ already achieves the SOTA generalization results(\autoref{tab:quantitative}), we observe the instance quality degrades, implying the model inevitably forgets the instance prior. 
As the spatial leaner $F$ is not directly learning the dataset layout, but learning from the given spatial guidance, we observe that whether the training scene layout has semantic meaning is not very important for \name.
In the other hand, training on the random scene composed of instances from the whole instance 3D dataset further improve \name\ generalization ability.

\noindent{\textbf{Collision-free Random Layout.}} 
We generate our synthetic scenes via sampling high‑quality 3D instances from a diverse 3D asset dataset(e.g., Objaverse~\cite{deitke2023objaverse}), randomly places them with a collision‑free mechanism to reduce severe occlusions, and enforces a common spatial relation such as \emph{right}, \emph{left}, \emph{front}, \emph{back}, and \emph{on the top of}.
The scene layout is purely non‑semantic meaningful composition of instance; only geometric and basic physical plausibility constraints are applied. 
By training on this dataset, \name\ learns general spatial reasoning while remaining agnostic to category semantics.
We provide our random layout generation details in \textit{appendix}.

\subsection{Training}
\label{subsection:training}
The training target is conditioned rectified flow methods:
\begin{equation}
    \mathcal{L}_{\mathrm{CFM}}(\theta) = \mathbb{E}_{t,x_0,\epsilon} \left\| v_{\theta}(x,t) - (\epsilon - x_0) \right\|_2^2,
\end{equation}
where $v_{\theta}$ is the sparse structure neural network, $x(t) = (1-t)x_0 + t\epsilon$, and $\epsilon$ is the noise at time step $t$.

\section{Experiment}

\begin{table*}[t]
\centering
\caption{Comparison on evaluation datasets. 
CD: Chamfer Distance; F-Score threshold $\tau{=}0.1$.
S = scene-level; O = object-level; IoU-B = volumetric IoU of scene bounding boxes.
Best numbers are in \textbf{bold}. PartCrafter is only compared on scene-level performance.
}
\label{tab:quantitative}

\small
\setlength{\tabcolsep}{3pt} 

\begin{tabular*}{\textwidth}{@{\extracolsep{\fill}}%
  l
  S[table-format=1.3] S[table-format=2.2] S[table-format=1.3] S[table-format=2.2] S[table-format=1.3]
  S[table-format=1.3] S[table-format=2.2] S[table-format=1.3] S[table-format=2.2] S[table-format=1.3]
  l@{}}
\toprule
\multirow{2}{*}{Method}
  & \multicolumn{5}{c}{\textbf{3D-FRONT}}
  & \multicolumn{5}{c}{\shortstack{\textbf{BlendSwap} \& \textbf{Scenethesis}}}
  & \multirow{2}{*}{Runtime$\downarrow$} \\
\cmidrule(lr){2-6}\cmidrule(lr){7-11}
  & {CD-S$\downarrow$} & {F-Score-S$\uparrow$} & {CD-O$\downarrow$} & {F-Score-O$\uparrow$} & {IoU-B$\uparrow$}
  & {CD-S$\downarrow$} & {F-Score-S$\uparrow$} & {CD-O$\downarrow$} & {F-Score-O$\uparrow$} & {IoU-B$\uparrow$} & \\
\midrule          
Gen3DSR
& {0.2587} & {42.31} & {0.0697} & {57.22} & {0.4838} 
& {0.1429} & {45.43} & {0.0722} & {53.45} & {0.4736}
& {179.0\ s} \\
PartCrafter  
& {0.0586} & {81.03} & {-} & {-} & {0.7626} 
& {0.0609} & {66.56} & {-} & {-} & {0.5819} 
& \tb{7.2\,s} \\
SceneGen     
& {0.1432} & {54.70} & {0.0353} & {77.95} & {0.5295} 
& {0.1161} & {49.94} & {0.0852} & {65.66} & {0.4669} 
& {26.0\,s} \\
MIDI    
& {0.0175} & {90.08} & {0.0877} & {70.10} & {0.8596} 
& {0.0212} & {83.13} & {0.1884} & {50.84} & {0.7412} 
&  42.5\,s \\
\rowcolor{black!03}
\textbf{Ours} 
& \tb{0.0148} & \tb{93.50} & \tb{0.0207} & \tb{84.28} & \tb{0.8762} 
& \tb{0.0059} & \tb{94.26} & \tb{0.0503} & \tb{72.39} & \tb{0.8568}
& {15.51\,s} \\
\bottomrule
\end{tabular*}
\end{table*}

\begin{table*}[t]
\centering
\small
\caption{
Comparison on \textbf{3D-FRONT} and \textbf{BlendSwap \& Scenethesis}. 
CD: Chamfer Distance; F-Score threshold $\tau{=}0.1$. 
\emph{S} = scene-level, \emph{O} = object-level; IoU-B = volumetric IoU of scene bounding boxes. Best numbers are \textbf{bold}.}
\setlength{\tabcolsep}{4pt} 
\newcommand{\best}[1]{\bfseries #1}

\begin{tabular*}{\textwidth}{@{\extracolsep{\fill}} 
  l
  S[table-format=1.3] S[table-format=2.2] S[table-format=1.3] S[table-format=2.2] S[table-format=1.3]
  S[table-format=1.3] S[table-format=2.2] S[table-format=1.3] S[table-format=2.2] S[table-format=1.3]
  l @{}}
\toprule
\multirow{2}{*}{Training dataset}
  & \multicolumn{5}{c}{\textbf{3D-FRONT}}
  & \multicolumn{5}{c}{\textbf{BlendSwap \& Scenethesis}}\\
\cmidrule(lr){2-6}\cmidrule(lr){7-11}
  & {CD-S$\downarrow$} & {F-Score-S$\uparrow$} & {CD-O$\downarrow$} & {F-Score-O$\uparrow$} & {IoU-B$\uparrow$}
  & {CD-S$\downarrow$} & {F-Score-S$\uparrow$} & {CD-O$\downarrow$} & {F-Score-O$\uparrow$} & {IoU-B$\uparrow$} \\
\midrule

3D-FT (25K)
& \tb{0.0137} & \tb{93.77} & {0.0278} & {81.34} & \tb{0.8792} 
& {0.0118} & {90.79} & {0.0585} & {68.87} & {0.8222} \\

Rand-15K
& {0.0496} & {79.96} & {0.0932} & {55.01} & {0.7729} 
& {0.0081} & {92.67} & {0.0698} & {67.36} & {0.8445} \\

Rand-25K
& {0.0406} & {81.39} & {0.0402} & {74.76} & {0.7783} 
& {0.0075} & {93.60} & {0.0580} & {70.18} & {0.8471} \\

3D-FT+Rand-15K
& {0.0148} & {93.50} & \tb{0.0207} & \tb{84.28} & {0.8762} 
& \tb{0.0059} & \tb{94.26} & \tb{0.0503} & \tb{72.39} & \tb{0.8568} \\

\bottomrule
\end{tabular*}
\label{tab:statistic}
\end{table*}

\subsection{Experiment Setting}

\noindent\textbf{Baseline.}
We compare the representative interactive 3D scene generation methods including MIDI~\cite{huang2025midi}, SceneGen~\cite{meng2025scenegen}, Partcrafter~\cite{lin2025partcrafter}, and Gen3DSR~\cite{ardelean2024gen3dsr}. 

For all baselines, we use the same scene RGB image and instance masks as inputs, except PartCrafter, which does not support mask control; for PartCrafter we provide the scene RGB plus the target instance count. 
Because PartCrafter lacks texture rendering, we evaluate it only on scene-level geometry and layout quality. For visual quality comparisons, we texture MIDI’s generated scenes using MV-Adapter~\cite{huang2025mv}.

\noindent\textbf{Metric.}
We evaluate scenes based on geometry quality and layout accuracy.
For geometry quality, we convert the generated assets to point clouds and rigidly align the
prediction to the ground truth using customized robust ICP~(details in \textit{appendix}). 

After alignment, we then report Chamfer Distance (CD) and F\mbox{-}Score (threshold
$\tau{=}0.1$) at two levels: (i) \emph{scene level} on the union of all points, and
(ii) \emph{object level} by computing the metrics per matched instance and averaging.

To evaluate scene layout, we compute matched volumetric IoU between the axis-aligned bounding boxes (AABBs) of predicted and ground-truth scenes.
This metric captures the overall objects size, position, and relative placement. 

We also report the average inference time per instance.

\noindent\textbf{Evaluation Dataset.}
We evaluate all methods on \textbf{synthetic} data and \textbf{real-world/stylized} scenes.
i) \textit{Synthetic (in-domain).}
Following MIDI, we adopt the 3D-FRONT split and use the same test scenes, filtering out renders with fully occluded instance masks; this yields \(\sim\)860 test scenes.
ii) \textit{Synthetic (out-of-domain).}
To assess generalization to novel layouts and object sets, we additionally evaluate on scenes from BlendSwap~\cite{azinovic2022neural} and Scenethesis~\cite{ling2025scenethesis}, which yield 26 test scenes including small on large relations and outdoor settings.
iii) \textit{Real-world /stylized.}
For in-the-wild evaluation without ground truth, we test on images from DL3DV-140~\cite{ling2024dl3dv}, Gen3DSR~\cite{ardelean2024gen3dsr}, ScanNet++~\cite{yeshwanth2023scannet++}, and stylized scenes. Then, we report qualitative comparisons.

\begin{figure*}[t]
    \centering
    \includegraphics[width=\linewidth]{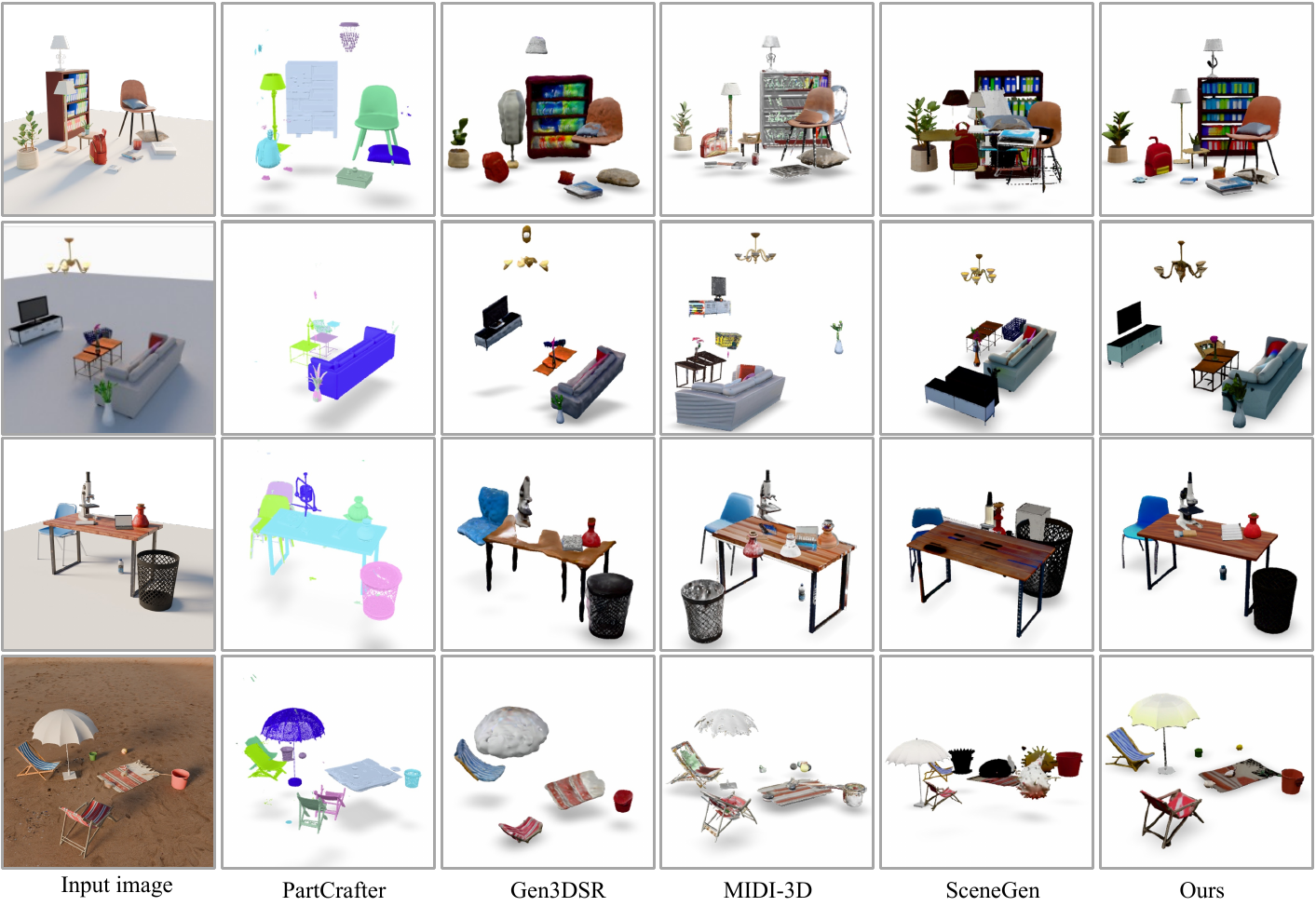}
    \caption{Qualitative comparison with all baselines on synthetic scenes. We might slightly rotate the view to better illustrate the error patterns. More details can be found in \textit{Appendix}.} 
    \label{fig:qual-synth}
\end{figure*}

\subsection{Quantitative Results}
\label{subsec:quantitative_result}

\noindent\autoref{tab:quantitative} reports quantitative comparisons on synthetic datasets, including 3D-FRONT~\cite{fu20213d},  BlendSwap~\cite{azinovic2022neural}, and Scenethesis~\cite{ling2025scenethesis}. 3D-FRONT dataset only contains bedroom/living room scenarios, while Blendswap and Scenethesis contain more diverse layout.
Our method, \name, achieves the best performance among the
state-of-the-art methods across all evaluated metrics without incurring much time consumption.

\noindent\textbf{Geometry quality.}
\textit{Object-level.}
On 3D‑FRONT test scenes (ID), our method attains the \emph{lowest} object‑level CD and the \emph{highest} F‑score compared with
MIDI~\cite{huang2025midi}, SceneGen~\cite{meng2025scenegen}, PartCrafter~\cite{lin2025partcrafter},
and Gen3DSR~\cite{ardelean2024gen3dsr}, indicating higher‑fidelity per‑object geometry.
Crucially, on out-of-domain (OOD) benchmarks (BlendSwap, Scenethesis), our object‑level scores remain \emph{comparable to ID}
and still exceed all baselines by a clear margin, evidencing robust generalization.
\textit{Scene-level.}
Across both ID and OOD settings, our method also achieves \emph{lowest} scene‑level CD and \emph{highest} F‑score than the baselines, reflecting more accurate and more complete whole‑scene geometry.
We attribute these gains to:
i) \textbf{reprogrammed a pre-trained instance prior.} A pre-trained instance generator supplies strong single‑object shape priors, which is an advantage over methods
rely primarily on reconstruction from limited data such as Gen3DSR;
ii) \textbf{model‑centric supervision with shared, view‑centric scene context}. 
Per‑instance generation conditioned on shared, view‑centric
scene tokens keeps shapes near canonical while aligning pose and contact, reducing fused‑mesh artifacts and “shape bending” common in
scene dataset‑supervised methods such as MIDI, SceneGen, PartCrafter.
Unlike all baselines, which degrade substantially on OOD scenes, our object‑ and scene‑level metrics on BlendSwap and Scenethesis remain close to our ID scene performance, underscoring strong generalization.

\noindent\textbf{Layout accuracy.}
Our method attains the highest \emph{Volumetric IoU} across all baselines, indicating more accurate object placement in
position, scale, and orientation and global scene geometry consistency. 
Gains are largest on \emph{BlendSwap}/\emph{Scenethesis}. 

Our results suggest stronger modeling of support and proximity for various layout.
The metric result also confirm the advancement of \name\ from scene and instance geometry quality. 
The instance semantic and spatial relation disentangled formulation improves scene‑level fidelity without additional scene annotations.

\noindent\textbf{Efficiency.}
We measure end‑to‑end per instance inference time on a single H100 GPU for all approaches. Our method completes a instance in 15.51s, although slower than PartCrafer (7.2 s) while delivering higher geometry and layout quality, and faster than the other baselines in our study. For fairness, MIDI inference time includes per instance texture rendering.
We attribute our latency to a feed‑forward design leveraging the pre-trained instance prior, which avoids retrieval or iterative layout optimization.

\begin{figure*}[t]
    \centering
    \includegraphics[width=\linewidth]{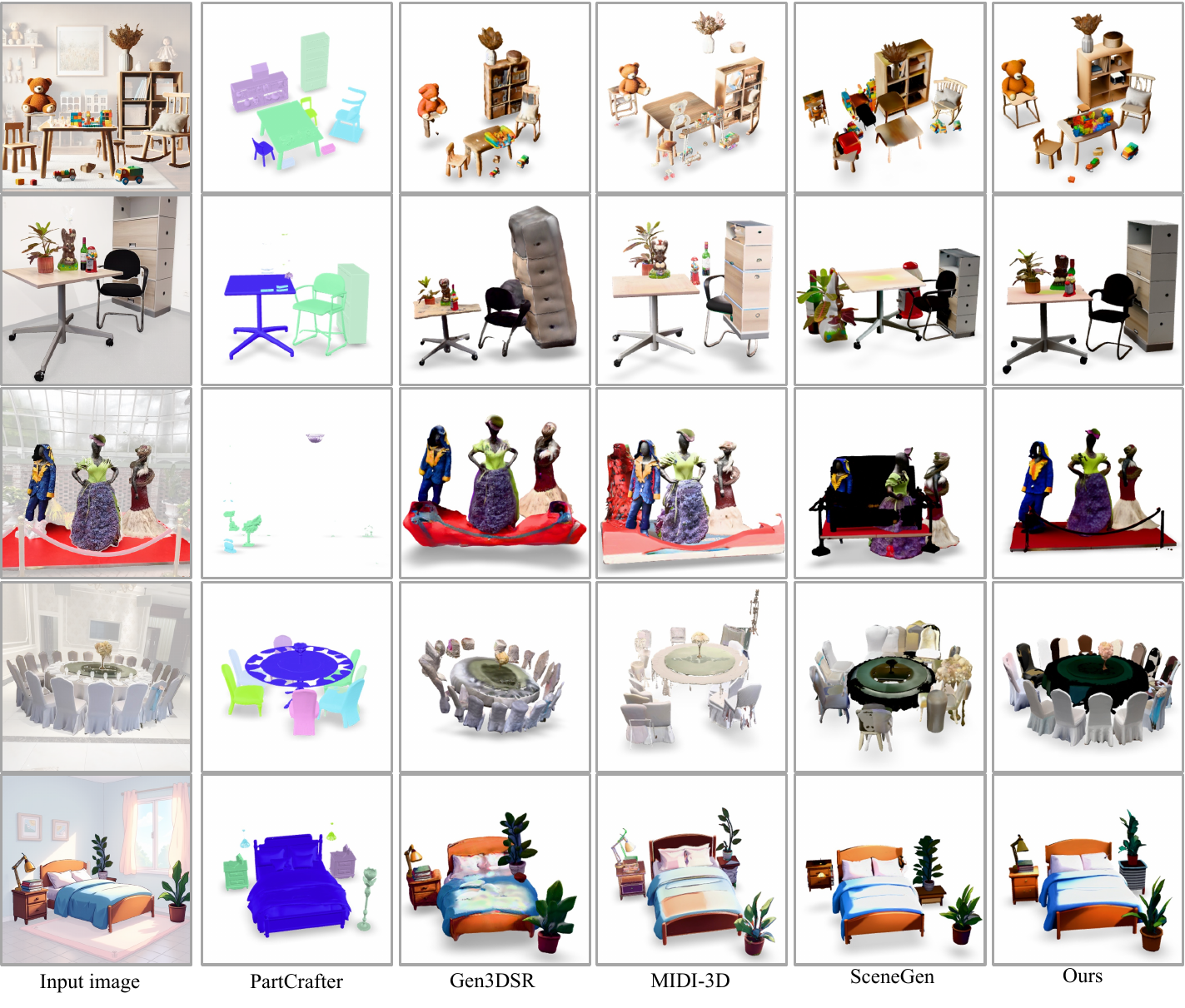}
    \caption{Qualitative comparison with scenes in-the -wild. We compare the generalization ability of different methods in different style and various spatial relations. To better illustrate the error pattern, we might slightly rotate the view.}
    \label{fig:qual-real}
\end{figure*}

\begin{table*}[t]
\newcommand{\cds}{CD-S$\downarrow$}
\newcommand{\fss}{F-Score-S$\uparrow$}
\newcommand{\cdo}{CD-O$\downarrow$}
\newcommand{\fso}{F-Score-O$\uparrow$}
\newcommand{\ious}{IoU-S$\uparrow$}
\centering
\small
\setlength{\tabcolsep}{3pt}
\renewcommand{\arraystretch}{1.05}
\caption{Ablation on \name's components. SCA: scene context attention, VC: view centric, NS: non-semantic scenes. }
\label{tab:ablation}
\begin{tabular*}{\textwidth}{@{\extracolsep{\fill}}%
  c c c 
  S[table-format=1.3] S[table-format=2.2] S[table-format=1.3] S[table-format=2.2] S[table-format=1.3] 
  S[table-format=1.3] S[table-format=2.2] S[table-format=1.3] S[table-format=2.2] S[table-format=1.3] 
  @{}}
\toprule
\multicolumn{3}{c}{\textbf{Examine component}} &
\multicolumn{5}{c}{\textbf{3D-FRONT}} &
\multicolumn{5}{c}{\textbf{BlendSwap \& Scenethesis}} \\
\cmidrule(lr){1-3}\cmidrule(lr){4-8}\cmidrule(lr){9-13}
\textbf{SCA} & \textbf{VC} & \textbf{NS} &
{\cds} & {\fss} & {\cdo} & {\fso} & {\ious} &
{\cds} & {\fss} & {\cdo} & {\fso} & {\ious} \\
\midrule
\checkmark & \ding{55} & \ding{55} 
& {0.0163} & {93.69} & {0.0286} & {80.12} & {0.8598} 
& {0.0351} & {79.12} & {0.0829} & {63.16} & {0.7557} 
\\
\checkmark & \checkmark & \ding{55} 
& {0.0137} & {93.77} & {0.0278} & {81.34} & {0.8792} 
& {0.0118} & {90.79} & {0.0585} & {68.87} & {0.8222} \\
\checkmark & \checkmark & \checkmark 
& {0.0148} & {93.50} & {0.0207} & {84.28} & {0.8762} 
& {0.0059} & {94.26} & {0.0503} & {72.39} & {0.8568} \\
\bottomrule
\end{tabular*}
\end{table*}

\subsection{Qualitative results}

We qualitatively compare our approach with MIDI, SceneGen, PartCrafter, and Gen3DSR in two settings: 
(i) \textbf{synthetic scenes} and 
(ii) \textbf{real‑world and stylized inputs}. 

All methods receive the same single input image.

\noindent\textbf{Synthetic scenes}
~\autoref{fig:qual-synth} contrasts our results on 3D‑FRONT‑style scenes and more complicated layouts from BlendSwap and Scenethesis.
\textbf{Instance quality.} Our method produces \emph{clean, well‑separated} instance meshes with sharp thin parts (e.g., shelf, sofa, umbrella in the figure) and minimal artifacts, aligning with the lower object‑level CD and higher F‑score in Sec.~\ref{subsec:quantitative_result}. 
Baseline methods such as MIDI and SceneGen often exhibit \emph{duplicate placements}, \emph{uncleaned mesh}, or \emph{over‑smoothed geometries} that blur instance boundaries.
\textbf{Spatial relations.} Our scenes preserve \emph{support} and \emph{proximity} (e.g., small‑on‑large relations such as TV/ flower  on tables), maintain correct \emph{depth ordering} for partially \emph{behind/occluded} items, and avoid \emph{floating} or \emph{colliding} objects, which is common in baselines.
This is consistent with our higher Volumetric IoU and scene‑level F‑score. 
\textbf{Robustness on various layouts.} Gains are most visible on BlendSwap and Scenethesis testing scenes, which include multi‑scale clutter, small‑on‑large configurations, and outdoor layouts that are under‑represented in curated indoor datasets.

\noindent\textbf{Real‑world and stylized scenes}
We further evaluate on in‑the‑wild images from DL3DV-140, Gen3DSR, and ScanNet++ and on stylized/cartoon inputs as shown in ~\autoref{fig:qual-real}. 
Across these distributions, baselines are less robust to strong style shifts and complex layouts, frequently showing \emph{misaligned instances}, \emph{surface bleed} between nearby assets, and \emph{implausible supports} such as frequently collied objects. 
In contrast, our method maintains \textbf{consistent scale and uprightness}, \textbf{clear separation between instances}, and \textbf{plausible contacts} across indoor, outdoor, and stylized scenes. 

\noindent\textbf{Comparison with baselines.}
Existing feed‑forward or retrieval/assembly pipelines tend to either sacrifice per‑object fidelity or lose global coherence when generalize to novel layouts: duplicate objects, merged/entangled assets, collied and floating placements are common failure modes. 
Our results exhibit both \emph{higher per‑instance fidelity} and \emph{more coherent spatial relations}, which we attribute to 
i) \textbf{model‑centric supervision} from a reprogrammed pre‑trained instance prior (preserving object fidelity) 
and ii) \textbf{view‑centric shared scene context} that stabilizes relative pose, depth ordering, and support for each instance from a shared space during generation.
The \textit{Appendix} includes additional synthetic/real/stylized examples, multi‑view renderings for each scene, and videos for large scenes with complex layouts.

\subsection{Non-semantic random scene experiment}
In this section, we want to answer the \textbf{ question}: \emph{do non‑semantic, randomly composed scenes provide meaningful spatial supervision? }

\noindent\textbf{Setup.}
To answer the question, we conduct the experiment that train \name\ on four datasets:
i) \textbf{3D-FT}: 25K 3D‑FRONT scenes including bedroom/living‑room with dedicated scene annotations.
ii) \textbf{Rand-15K}: 15K non‑semantic scenes by randomly composing instances. It is generated by our synthetic data generation system.
iii) \textbf{Rand-25k}: same as Rand‑15K but scaled to 25K scenes.
iv)  \textbf{3D-FT+Rand-15K}: a mixture of 3D‑FT and Rand‑15K.

\noindent\textbf{Evaluation.}
The 3D‑FRONT test set is treated as \emph{in‑distribution} for model trained on 3D-FT. 
BlendSwap and Scenethesis contain more diverse layouts and are used as
\emph{out‑of‑distribution (OOD)} benchmarks.

\noindent\textbf{Result.}
~\autoref{tab:statistic} presents the experiment results. 
We observe that training solely on 3D‑FRONT (3D‑FT) yields the stronger in‑distribution layout (lowest CD‑S, highest IoU‑B) compered to purely trained on non-semantic scenes, but generalizes poorly to BlendSwap/Scenethesis (OOD). 
Interestingly, purely non‑semantic randomized training (Rand‑15k/25k) captures transferable spatial regularities and surpasses 3D‑FT on BlendSwap/Scenethesis scene‑level metrics (CD‑S↓, F‑S↑, IoU‑B↑), with performance improving as we scale from 15k to 25k scenes. 
The combined regimen (3D‑FT+Rand‑15k) preserves near‑optimal in-domain layout while further improving object geometry (lower CD‑O, higher F‑O) and achieving the best OOD results across all metrics. 

These findings indicate that 
i) \name\ is able to learn spatial knowledge from \emph{non‑semantic} scenes.
In particular, rather than relying on high-level scene semantics, pure geometric cues alone provide a strong supervision signal for spatial learning and reasoning. This observation suggests a broader scaling law: analogous to MegaSynth~\cite{jiang2025megasynth} in 3D reconstruction, \textbf{scaling synthetic, non-semantic layouts would provide richer spatial coverage that the model can exploit thus unlock large-scale interactive 3D scene generation}.
ii) Curated scene annotations remain useful for \emph{in‑distribution calibration},  \emph{synergizing training} with non-semantic scenes to yield the best overall performance.

\subsection{Ablation studies}
We evaluate the contribution of the main design choices in \name\ on 3D‑FRONT, BlendSwap, and Scenethesis.  
We study three components:
(i) \textbf{Scene‑context attention(SCA)}: shared scene-level tokens that contain spatial cues; 
(ii) \textbf{View‑centric(VC)}: the view‑aligned space; 
(iii) \textbf{Non‑semantic meaningful scenes (NS)}: randomized compositions from the Objaverse instances. 

We start from the full model (\emph{SCA + VC + NS}) and remove one component at a time.
As summarized in \autoref{tab:ablation}, we observe that
\textbf{every component matters.} Removing any of SCA, VC, NS consistently degrades geometry (CD$\uparrow$, F‑score$\downarrow$) and layout accuracy (IoU$\downarrow$) on all evalaution datasets, confirming their complementary roles in \name\ for generalize to novel layout.
\textbf{VC is critical for layout coherence.} Without view‑centric encoding, the model tends to over-fit on 3D-FTRONT dataset and has poor generalized results on Blendswap and Scenethesis. It also misaligns scene context across objects, yielding duplicated or ``blended'' instances and contact violations; this manifests as the largest IoU drop and a notable decline in scene‑level F‑score.
\textbf{NS improves instance generalization.}

Adding NS yields substantial gains in F‑score at the instance level and increment in spatial-relation (IoU), reflecting better surface completeness and fewer near‑surface errors for new scenes and inter‑object placement is driven chiefly by the view‑centric shared scene context.
We provide qualitative results for the ablation study in \textit{appendix}.
Overall, these results demonstrate that view‑centric scene context (VC+SC) is essential for coherent layouts and clean instance boundaries, while training on non‑semantic scene (NS) supplies the instance diversity needed to generalize beyond distributions.

\section{Conclusion}

\noindent{\textbf{Contribution.}}
We reprogram a pre‑trained 3D \emph{instance} generator as a scene‑level spatial learner. 
We define a \emph{view‑centric} scene space and distill spatial cues into shared \emph{scene tokens} that condition instance denoising, yielding a fully \emph{feed‑forward} formulation for interactive scene generation. 
This formulation replaces dataset‑bounded layout supervision with model’s own spatial prior, removing reliance on curated scene annotations and enabling learning spatial relations from \emph{non‑semantic} meaningful scenes. 
Our experiments indicate non semantic meaningful synthetic data improve geometry, layout, and generalization, pointing toward a foundation model for interactive 3D scene generation.

\noindent{\textbf{Limitations and future work.}}
Our method performs relatively poorly on tiny‑resolution inputs and in heavily occluded single‑view case (see \textit{Appendix}).
Future work: 1) improve model robustness with heavy occlusion augmentations, and to explore optional multi‑view conditioning; 2) Further investigate the scaling law of non-semantic random scene to further handle challenging in-the-wild layouts.

\clearpage
\setcounter{page}{1}
\maketitlesupplementary

\section{Overview}
\label{sec:rationale}

To better illustrate our method, the supplementary material is organized into two main parts: \textbf{Method} and \textbf{Experiments}. In addition, we provide an \textbf{HTML} page for improved qualitative visualization and comparison.

\paragraph{Method.}
\begin{itemize}
    \item \textbf{Scene-context attention.} We provide a mathematical derivation showing that when the scene and instance inputs are identical, the proposed scene-context attention (SCA) is equivalent to a standard self-attention layer.
    \item \textbf{Collision-free random layout.} We describe the details of our collision-free random layout generation procedure.
\end{itemize}

\paragraph{Experiment setting.}
\begin{itemize}
    \item Model implementation details.
    \item Training datasets.
    \item Metrics. We provide the full specification of our customized robust ICP metric and related details.
\end{itemize}

\paragraph{Qualitative results.}
\begin{itemize}
    \item Qualitative comparison with all baselines on synthetic scenes. We either (i) show interactive 3D scene comparisons on the web page, or (ii) include two representative rendered views per scene for each method.
    \item Qualitative comparison with all baselines on real-world and stylized images, following the same visualization protocol as above.
    \item Additional qualitative results for the \textbf{ablation} study.
    \item \textbf{Failure cases}. We report the failure cases when instance mask is small. The figure including the input image, the predicted scene, and the predicted low-quality instance examples.
\end{itemize}

\section{Method}
\paragraph{scene-context attention.}
In this section, we show the mathematic proof that when the $K_i/V_i$ is the same with the scene $K_{s}/V_{s}$, meaning when the scene and instance input is exact the same, the SCA gives equivalent result of self-attention layers.

Let $Q_i \in \mathbb{R}^{t \times d}$ be the (row-stacked) instance queries, $K_i, K_{\text{s}} \in \mathbb{R}^{n \times d}$ the keys, and $V_i, V_{\text{s}} \in \mathbb{R}^{n \times d_v}$ the values. We concatenate along the token dimension,
\[
\tilde{K}_i = [K_i;\,K_{\text{s}}] \in \mathbb{R}^{2n \times d}, 
\qquad 
\tilde{V}_i = [V_i;\,V_{\text{s}}] \in \mathbb{R}^{2n \times d_v},
\]
and define scene-context attention (SCA) by
\[
\mathrm{SCA}(Q_i,\tilde{K}_i,\tilde{V}_i)
= \softmax\!\Bigl(\tfrac{Q_i \tilde{K}_i^\top}{\sqrt{d}}\Bigr)\,\tilde{V}_i,
\]
where $\softmax(\cdot)$ is applied \emph{row-wise} across the key dimension. We show that when the scene and instance inputs coincide, i.e., $K_i = K_{\text{s}}$ and $V_i = V_{\text{s}}$, SCA reduces exactly to standard self-attention:
\[
\mathrm{SCA}(Q_i,\tilde{K}_i,\tilde{V}_i)
= \softmax\!\Bigl(\tfrac{Q_i K_{\text{s}}^\top}{\sqrt{d}}\Bigr) V_{\text{s}}.
\]

\noindent\textbf{Proposition.}
If $K_i = K_{\text{s}}$ and $V_i = V_{\text{s}}$, then
\[
\mathrm{SCA}(Q_i,[K_i;\,K_{\text{s}}],[V_i;\,V_{\text{s}}])
= \softmax\!\Bigl(\tfrac{Q_i K_{\text{s}}^\top}{\sqrt{d}}\Bigr) V_{\text{s}}.
\]

\noindent\textit{Proof.}
Under $K_i = K_{\text{s}}$ and $V_i = V_{\text{s}}$, write $K \coloneqq K_{\text{s}}$ and $V \coloneqq V_{\text{s}}$. Then $\tilde{K}_i = [K;\,K]$ and $\tilde{V}_i = [V;\,V]$. Let
\[
Z \coloneqq \frac{Q_i K^\top}{\sqrt{d}} \in \mathbb{R}^{t \times n}.
\]
By block structure,
\[
\frac{Q_i \tilde{K}_i^\top}{\sqrt{d}}
= \frac{Q_i [K;\,K]^\top}{\sqrt{d}}
= \bigl[\,Z \;\; Z\,\bigr] \in \mathbb{R}^{t \times 2n}.
\]
Consider any row $z \in \mathbb{R}^n$ of $Z$. The row-wise softmax over the concatenation $[z,z] \in \mathbb{R}^{2n}$ yields
\[
\softmax([z,z]) 
= \Bigl[\tfrac{1}{2}\softmax(z) \;\; \tfrac{1}{2}\softmax(z)\Bigr],
\]
because for each coordinate $j$, with $s \coloneqq \sum_{\ell=1}^n e^{z_\ell}$ we have
\[
\frac{e^{z_j}}{\sum_{\ell=1}^n e^{z_\ell} + \sum_{\ell=1}^n e^{z_\ell}}
= \frac{e^{z_j}}{2s}
= \tfrac{1}{2}\,\softmax(z)_j.
\]
Applying this row-wise to $[Z,Z]$ gives
\[
\softmax([Z,Z]) = \bigl[\,\tfrac{1}{2}S \;\; \tfrac{1}{2}S\,\bigr], \quad\text{where } 
\]
\[
S \coloneqq \softmax(Z) \in \mathbb{R}^{t \times n}.
\]
Therefore,

\begin{equation}
\begin{aligned}
\operatorname{softmax}([Z,Z])\,\tilde{V}_i
&= \bigl[\,\tfrac{1}{2}S \;\; \tfrac{1}{2}S\,\bigr]\,[V;\,V] \\
&= \tfrac{1}{2}SV + \tfrac{1}{2}SV \\
&= SV \\
&= \operatorname{softmax}\!\Bigl(\tfrac{Q_i K^\top}{\sqrt{d}}\Bigr)\,V.
\end{aligned}
\end{equation}
This is precisely the output of a standard self-attention layer evaluated on $(Q_i,K,V)$. \hfill$\blacksquare$

\paragraph{Collision-free Random Layout.} 
When we randomly create a layout, we first randomly sample $N$ instances from the instance object dataset (e.g. Objaverse). 
Then we generate collision-free layouts in a Poisson noise pattern by treating each object \(i\) as a 2D disc on the ground with radius \(r_i = s_i\,\hat{r}_i\), where \(\hat{r}_i\) is computed from the mesh x/z extents and \(s_i\) is the sampled scale. 
Centers are sampled with a variable‑radius Poisson‑disk routine that places larger radius first. 
We define a global clearance $\mathrm{gap}=\bar{r} \cdot s$, where $\bar{r}$ is the mean of ${r_i}$ and $s$ is a randomly sampled scaling factor that controls the layout density. 
A candidate center \(\mathbf{x}_i\) is accepted only if \(\|\mathbf{x}_i-\mathbf{x}_j\|_2 \ge r_i+r_j+\mathrm{gap}\) for all placed \(j\). We accelerate checks with a uniform grid of cell size \((\min_i r_i+\mathrm{gap})/\sqrt{2}\), retry up to a fixed budget, and expand the sampling region by 10\% if needed. 
When a table is present, it is placed first and included in the same exclusion process. 
For the stacked object, we slice the table mesh just below its top to extract the top polygon, use its centroid as the anchor, and choose a scale that fits inside based on the distance to the nearest edge, then place the stacked object on the table.

\section{Experiment}
\paragraph{Model implementation.}
All experiments are conducted on a cluster with 8 NVIDIA H100 GPUs.
We train \name\ for 130K steps using the AdamW optimizer with a learning rate of $5e^{-5}$ and a batch size of 8 per GPU.
During inference, we adopt 25 sampling steps with the classifier-free guidance set to ${\omega=3.0}$ for both the sparse structure generation and structured latents generation.

\paragraph{Training dataset}
The 3D-FRONT training dataset is the same as MIDI processed 3D-FRONT dataset, where there are roughly 24K different scenes. 
Using the random layout generation algorithm discussed above, randomly generate 15K and 25K scenes. Each scene has minimum 2 object and maximum 12 objects. 

For each scene, we use Blender to render 150 views that evenly distributed from all directions looking at the scene center. 
Then we transform each instance into view centric space, voxelize and encode each instance geometry using TRELLIS sparse structure encoder as ground truth. 
Note, in some view some object is fully occluded and not visible, we discard this render view to avoid the model memorize and hallucinate invisible objects.

\paragraph{Robust ICP.} 
Previous work MIDI~\cite{huang2025midi} uses ICP for alignment before calculating the metrics. 
We notice this ICP is not robust and leads to many bad alignment, making the metric calculation not reliable. 
Instead, we propose a robust ICP to make the metric results more reliable. 

The main reason ICP is not robust is it is very easy to get stuck into local minimum. To handle this challenge, we used several ways to escape local minimum. 

\paragraph{Initial transform search.}
We perform a yaw-sweep global initialization about a designated up axis
($a\in\{x,y,z\}$). For a candidate set of yaw angles (default $\{0,45,90,135,180,225,270,315\}$ degrees),
we rotate the source downsampled point cloud about $a$, and pre-score each angle using a
trimmed symmetric Chamfer distance with trim ratio $\tau{=}0.2$ on up to $2000$ sampled
points per cloud. 
We keep the top three yaw candidates by this pre-score and run a short seed ICP on the
downsampled clouds for each candidate using a point-to-point estimator (optionally with
isotropic scale if enabled). We select the initialization $T_0$ that minimizes
$\text{rmse} + \lambda(1-\text{fitness})$ with $\lambda$ set to the voxel size
($\lambda{=}v$). 

\paragraph{Coarse-to-fine.}

\emph{Shared normalization.} For numerical stability and consistency across stages, we
apply the same uniform normalization to source and target: let $c$ be the midpoint of the
axis-aligned bounding box over both clouds and $\sigma$ the maximum side length. We work
in normalized coordinates $x'=(x-c)/\sigma$ for the remainder of the registration.

\emph{Downsampling and normals.} We voxel downsample both normalized clouds with voxel
size $v=0.03$ (relative to the normalized extent) for the coarse stage, and estimate
normals on both downsampled and full-resolution clouds
when available.

\emph{Estimators.} The coarse stage uses a point-to-point objective; if isotropic scale is
enabled and supported, we estimate a global uniform scale jointly with the rigid motion.
The fine stage uses point-to-plane with a Tukey robust loss (scale $k=1.5v$); if normals
or robust losses are unavailable, we fall back to point-to-point.

\emph{Iteration budgets and thresholds.} We split the budget evenly with safeguards:
$T_{\mathrm{coarse}}=\max(10,\lfloor 0.5\,T\rfloor)$ and
$T_{\mathrm{fine}}=\max(10,T-T_{\mathrm{coarse}})$. Distance thresholds are $2.5v$ (coarse)
and $v$ for point-to-plane fine refinement (or $1.5v$ if point-to-point is used).

\emph{Execution.} Coarse ICP is run on the downsampled pair from $T_0$, yielding
$T_{\mathrm{coarse}}^\star$. If isotropic scale was estimated, we pre-apply
$T_{\mathrm{coarse}}^\star$ to the full-resolution source and run fine ICP from identity,
then compose the results; otherwise, we run fine ICP on the full-resolution pair
initialized with $T_{\mathrm{coarse}}^\star$.

\paragraph{Pose projection and validation.}
We project the
$3{\times}3$ rotation block to the nearest element of $\mathrm{SO}(3)$ via SVD. If
reflections are disallowed, we enforce a positive determinant. We then perform sanity
checks on the transform in normalized space (finite entries, reasonable determinant when
rigid, translation magnitude not excessive). On detection of anomalies we fall back to the
coarse solution or identity, whichever is valid.

\paragraph{Practical notes.}
As different methods generate in different space with various scale, this introduce an subtle bias. Even if we align the space by min/max values, the scene size potentially still affects the results. 
For example, rotating an extreme long scene a little bit potentially has different scene scale after normalization. 
So we apply the robust ICP alignment in this way, say we have two space: (i). min/max normalize space: just scale the scene space by the default output range, e.g. TRELLIS outputs into [-0.5, 0.5], then we just scale the scene isotropically by 2; (ii). AABB recentered normliaze: we move the scene into zero center and then normalize AABB min/max into [-1.0, 1.0]. 
We apply robust ICP and obtain $transform_1$ and $transform_2$. We then pick the best metric results from the two transformations.

\begin{figure*}[t]
    \centering
    \includegraphics[width=0.9\linewidth]{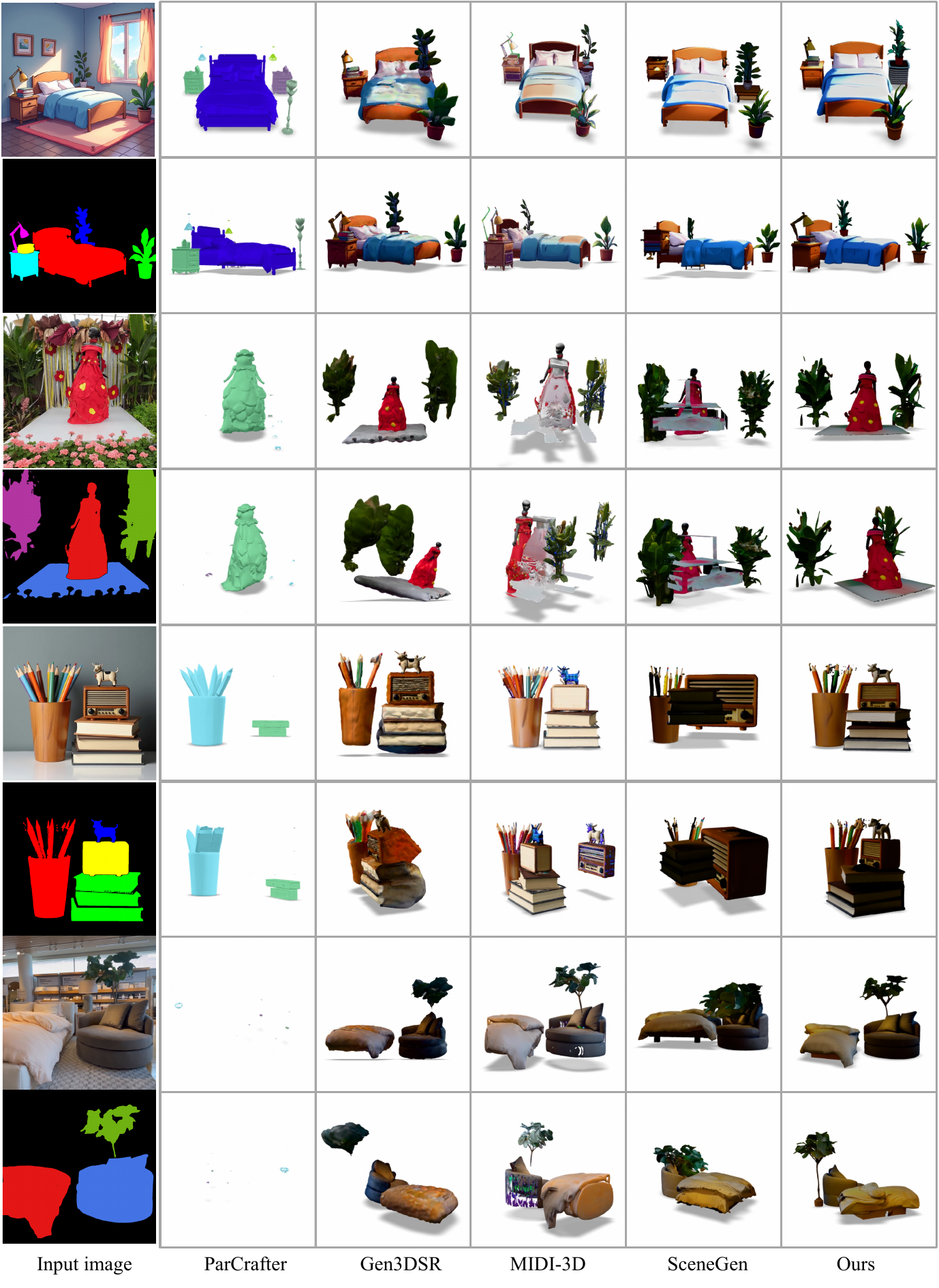}
    \caption{We present the multi-view evaluation results using real-world /stylized image as input (example 1). The testing scenes contain various layouts including \textit{front, back, right, left, small on large, behind}, and etc. Baselines includes PartCfrater~\cite{lin2025partcrafter}, Gen3DSR~\cite{ardelean2024gen3dsr}, MIDI-3D~\cite{huang2025midi}, and SceneGen~\cite{meng2025scenegen}}
    \label{fig:real-world-qualitative1}
\end{figure*}

\begin{figure*}[t]
    \centering
    \includegraphics[width=0.9\linewidth]{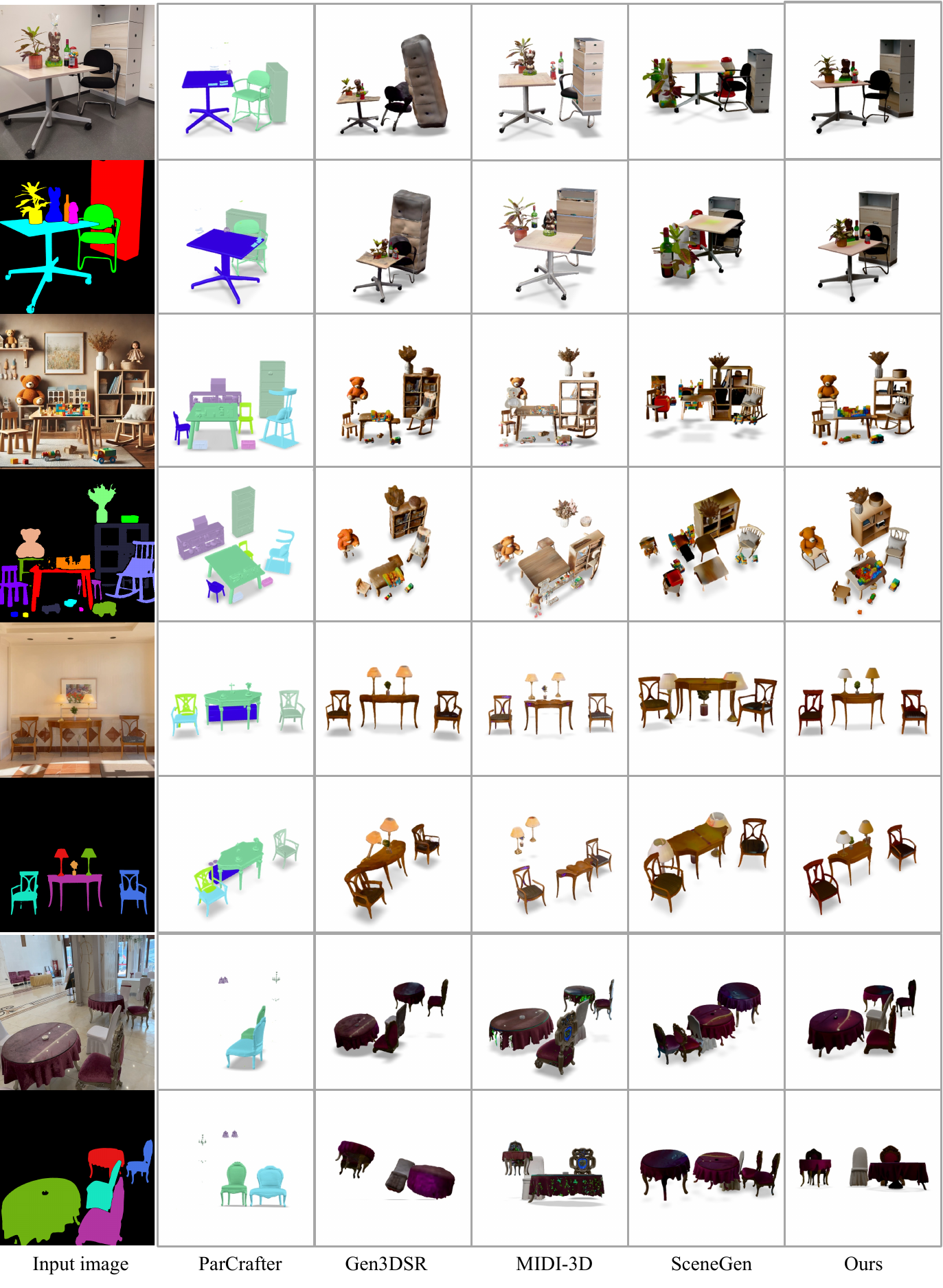}
    \caption{We present the multi-view evaluation results using real-world /stylized image as input (example 2). The testing scenes contain various layouts including \textit{front, back, right, left, small on large, behind}, and etc. Baselines includes PartCfrater~\cite{lin2025partcrafter}, Gen3DSR~\cite{ardelean2024gen3dsr}, MIDI-3D~\cite{huang2025midi}, and SceneGen~\cite{meng2025scenegen}}
    \label{fig:real-world-qualitative2}
\end{figure*}

\section{Qualitative results.}
\begin{figure*}[t]
    \centering
    \includegraphics[width=0.9\linewidth]{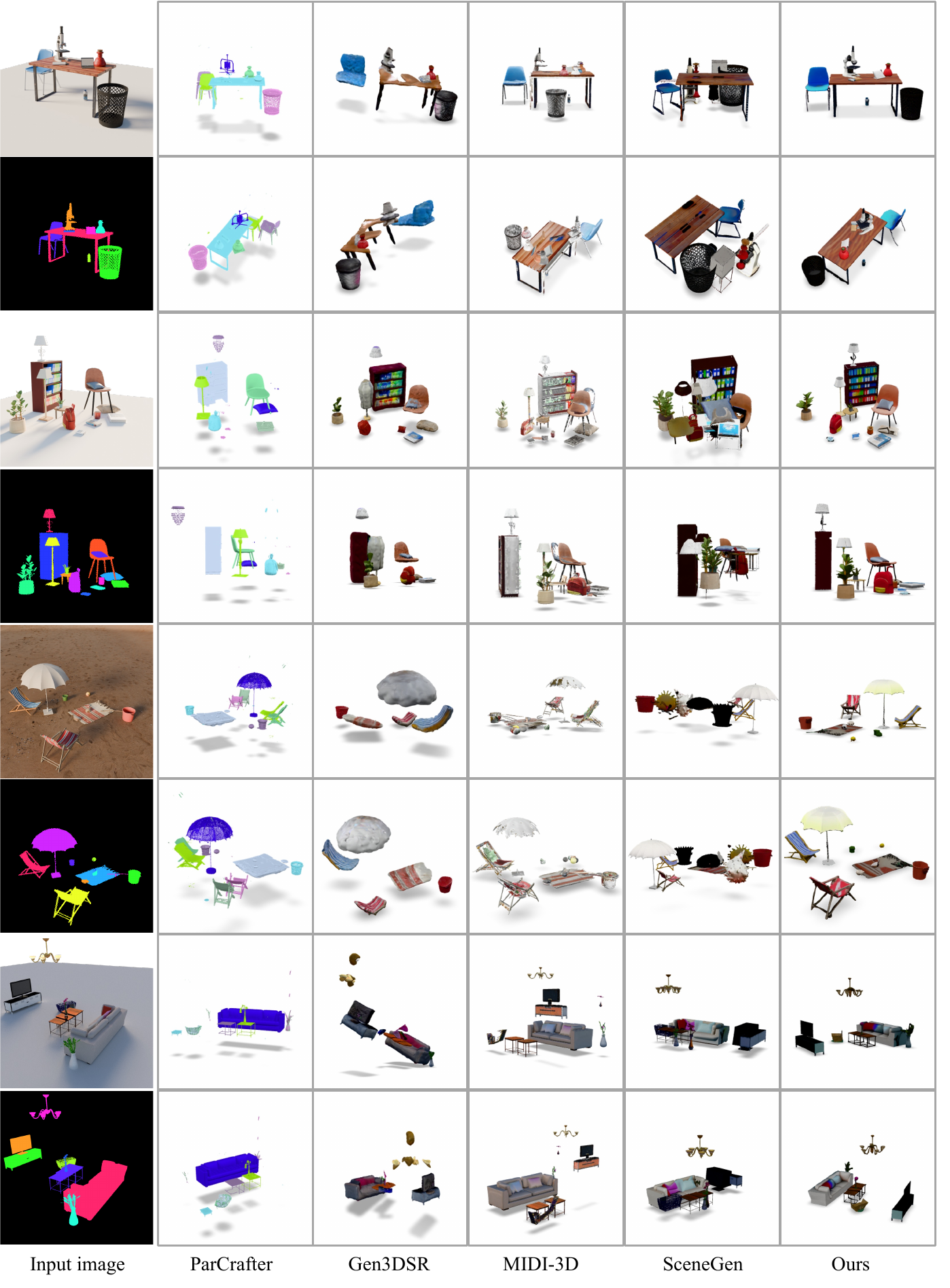}
    \caption{We present the multi-view evaluation results using synthetic scenes as input. The testing scenes contain various layouts including \textit{front, back, right, left, small on large, behind}, and etc. Baselines includes PartCfrater~\cite{lin2025partcrafter}, Gen3DSR~\cite{ardelean2024gen3dsr}, MIDI-3D~\cite{huang2025midi}, and SceneGen~\cite{meng2025scenegen}}
    \label{fig:real-world-synthetic}
\end{figure*}

\begin{figure*}[t]
    \centering
    \includegraphics[width=0.95\linewidth]{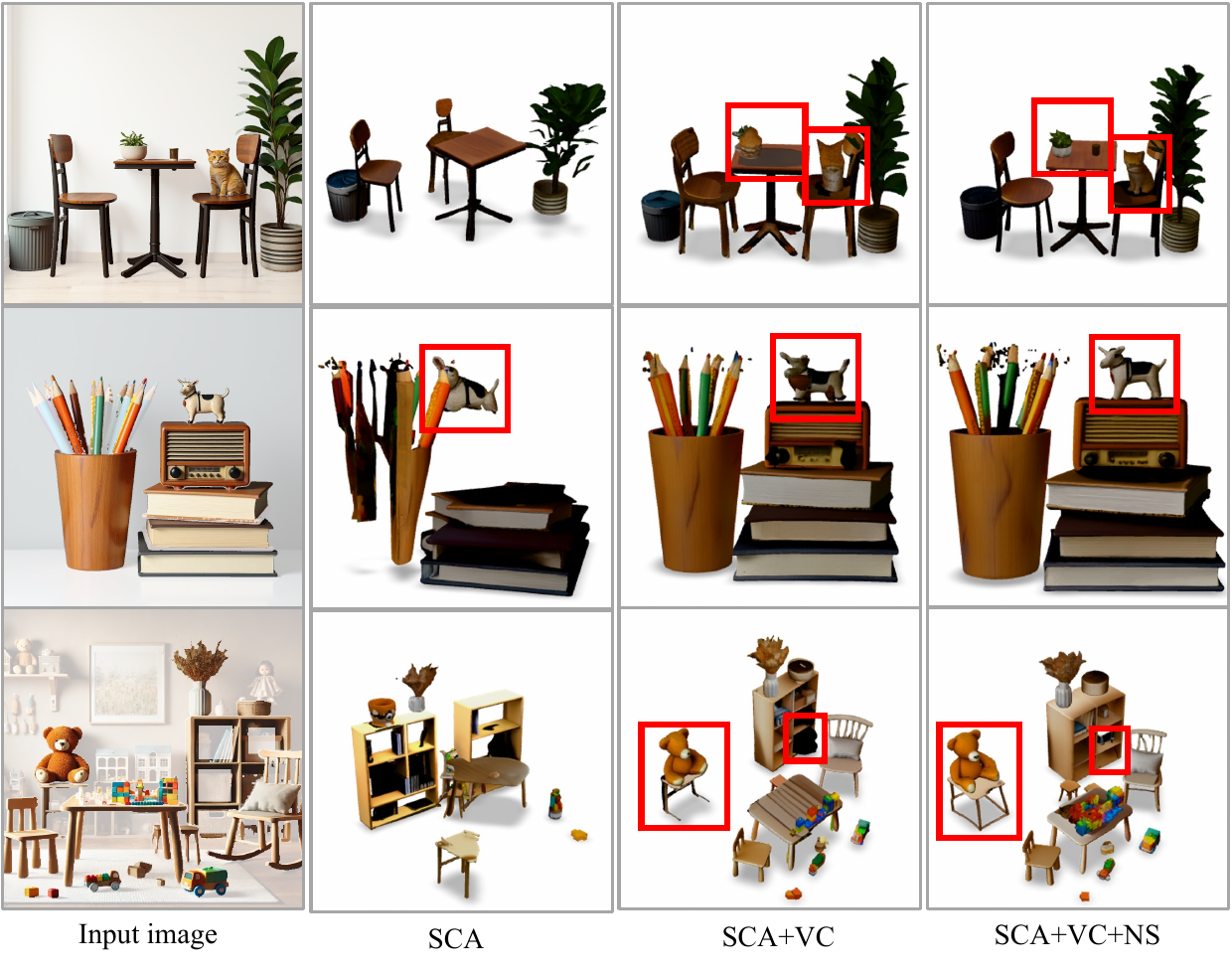}
    \caption{Ablation study on scene-context attention (SCA), view-centric space (VC), and non-semantic scenes (NS). With only SCA, the model often fails to maintain a coherent global layout and produces frequent object collisions. Adding the VC component substantially improves the overall arrangement of objects, but instance quality remains limited, as seen in the distorted cat, cow, and the chair under the toy bear. Incorporating NS further enhances both instance fidelity and global layout coherence.}
    \label{fig:ablation}
\end{figure*}

\begin{figure*}[t]
    \centering
    \includegraphics[width=0.95\linewidth]{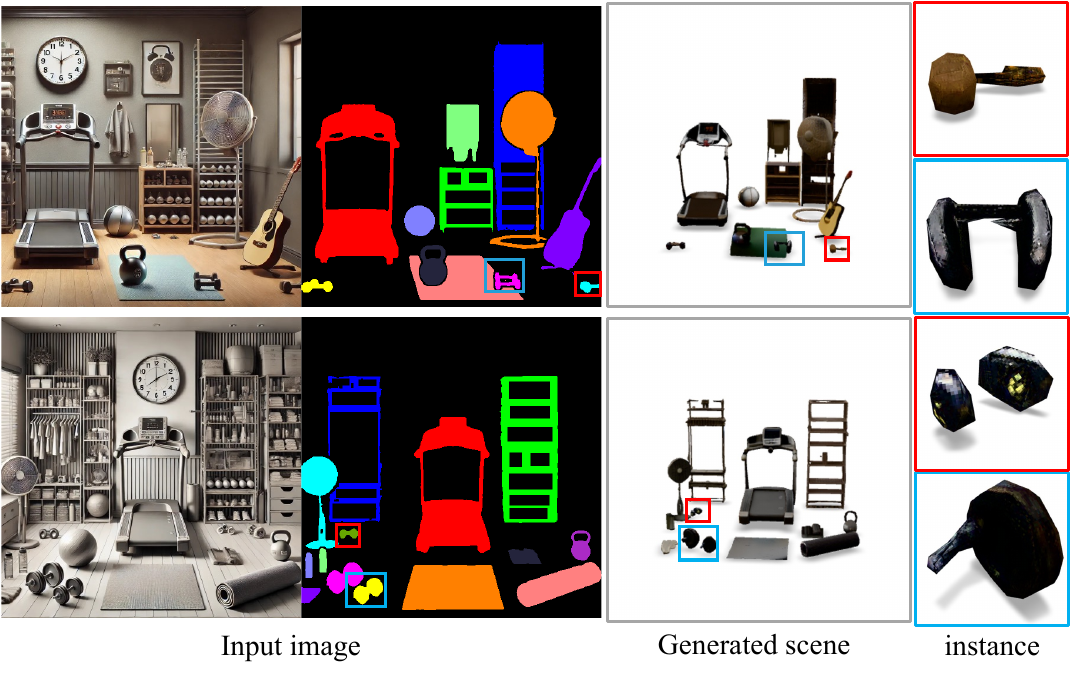}
    \caption{instance quality is bad when the input instance mask is small in the image.}
    \label{fig:fail}
\end{figure*}

We present the qualitative results for the following items:
\begin{itemize}
    \item \autoref{fig:real-world-qualitative1} and \autoref{fig:real-world-qualitative2} present multi-view qualitative comparison between \name\ and all baselines using real-world /stylized images as input.
    \item \autoref{fig:real-world-synthetic} presents multi-view qualitative comparison between \name\ and all baselines using synthetic scene images as input.
    \item The visualization of ablation study can be found in~\autoref{fig:ablation}, where adding each component (SCA, VC, and NS) in \name\ would significantly improve the layout coherent and instance quality.
    \item We also present the failure cases in~\autoref{fig:fail}. As shown in the figure, instance quality is bad when the input instance mask is small in the image.
    
\end{itemize}

{
    \small
    \bibliographystyle{ieeenat_fullname}
    \bibliography{main}
}

\end{document}